\documentclass[conference]{IEEEtran}
\IEEEoverridecommandlockouts
\usepackage{cite}
\usepackage{amsmath,amssymb,amsfonts}
\usepackage{algorithmic}
\usepackage{graphicx}
\usepackage{textcomp}
\usepackage{xcolor}
\usepackage{siunitx}
\usepackage{multirow}
\usepackage{subcaption}

\def\BibTeX{{\rm B\kern-.05em{\sc i\kern-.025em b}\kern-.08em
    T\kern-.1667em\lower.7ex\hbox{E}\kern-.125emX}}
\begin{document}

\title{Vital Sign Forecasting for Sepsis Patients in ICUs
\thanks{We would like to acknowledge Spass Inc., South Korea for funding this research.}
}

\author{\IEEEauthorblockN{Anubhav Bhatti$^{1}$, Yuwei Liu$^{1, 2}$, Chen Dan$^{1,2}$, Bingjie Shen$^{1,2}$, San Lee$^{1}$, Yonghwan Kim$^{3}$, Jang Yong Kim$^{4}$}
\IEEEauthorblockA{\textit{$^{1}$AI Engineering Team, SpassMed Inc., $^{2}$University of Toronto}, Canada, \\
\textit{$^{3}$Spass Inc., $^{4}$St. Mary’s Hospital}, South Korea\\
(anubhav.bhatti, yuwei.liu, chen.dan, bingjie.shen, sanlee)@spassmed.ca, kyh@spass.ai, vasculakim@catholic.ac.kr}
}

\maketitle
\thispagestyle{plain}
\pagestyle{plain}

\begin{abstract}
Sepsis and septic shock are a critical medical condition affecting millions globally, with a substantial mortality rate. This paper uses state-of-the-art deep learning (DL) architectures to introduce a multi-step forecasting system to predict vital signs indicative of septic shock progression in Intensive Care Units (ICUs). Our approach utilizes a short window of historical vital sign data to forecast future physiological conditions. We introduce a DL-based vital sign forecasting system that predicts up to 3 hours of future vital signs from 6 hours of past data. We further adopt the DILATE loss function to capture better the shape and temporal dynamics of vital signs, which are critical for clinical decision-making. We compare three DL models, N-BEATS, N-HiTS, and Temporal Fusion Transformer (TFT), using the publicly available eICU Collaborative Research Database (eICU-CRD), highlighting their forecasting capabilities in a critical care setting. We evaluate the performance of our models using mean squared error (MSE) and dynamic time warping (DTW) metrics. Our findings show that while TFT excels in capturing overall trends, N-HiTS is superior in retaining short-term fluctuations within a predefined range. This paper demonstrates the potential of deep learning in transforming the monitoring systems in ICUs, potentially leading to significant improvements in patient care and outcomes by accurately forecasting vital signs to assist healthcare providers in detecting early signs of physiological instability and anticipating septic shock. 
\end{abstract}


\section{Introduction}
Sepsis is a severe medical condition that can significantly threaten one's life. It occurs when the body's immune system responds to an infection by releasing chemicals in the bloodstream, leading to inflammation and potential damage to tissues and organs \cite{kim2019sepsis,calvert2016computational,wacker2013procalcitonin}. In North America alone, around 1.7 million people are affected by sepsis yearly, and roughly 270,000 cases result in death \cite{ferreras2015implementacion,wacker2013procalcitonin}. Globally, sepsis claims the lives of approximately 6 million out of the 30 million who develop the condition \cite{hall2011inpatient}. Monitoring vital signs is crucial in healthcare settings to detect early signs of physiological deterioration and take necessary actions to improve the patient outcomes \cite{kumar2006duration}. 
Traditional scoring frameworks such as the Acute Physiology, Age, Chronic Health Evaluation (APACHE) \cite{knaus1981apache}, Simplified Acute Physiology Score (SAPS) \cite{le1984simplified}, and Sequential Organ Failure Assessment (SOFA) \cite{jones2009sequential}, and qSOFA \cite{marik2017sirs}, are predicated upon physiological metrics to ascertain the severity of sepsis in patients. 
Nevertheless, these systems are not configured to detect sepsis or septic shock at an early stage, highlighting the urgent necessity to create more sophisticated monitoring methods that can enable swift medical action. Moreover, the need for clinical input to calculate SOFA and qSOFA scores makes them impractical for real-time detection systems, limiting their effectiveness in proactive sepsis management.

Recent studies have shown that machine learning and deep learning-based forecasting models have great potential in classification and forecasting time series data \cite{kane2014comparison,zhang2003time,wang2023long,behinaein2021transformer,bhatti2021attentive}. These models have demonstrated the ability to learn complex patterns and relationships within time series data and enable accurate predictions of future values. In recent years, several state-of-the-art deep learning forecasting models, including N-BEATS \cite{oreshkin2020nbeats}, N-HiTS \cite{challu2023nhits}, and Temporal Fusion Transformer (TFT) \cite{lim2019temporal}, have emerged as promising approaches for time series forecasting tasks. These deep learning-based architectures have demonstrated superior performance in various applications, including energy consumption forecasting, financial time series prediction, and weather forecasting, and have enabledtransfer learning in time series forecasting. However, these techniques have not been applied in forecasting vital signs for critical care patients.

In this work, we introduce a deep learning-based multi-step forecasting system for forecasting vital signs using previously observed vital signs of patients in the Intensive Care Units (ICUs) suffering from sepsis in critical care. Accurately forecasting the vital signs has the potential to assist healthcare providers in clinical settings such as ICUs in detecting early signs of physiological instability and anticipating changes in a patient's condition e.g., septic shock that is defined as a subset of sepsis. Our contribution in this paper can be summarized as follows: \noindent \textbf{(1)} We introduce a deep learning vital sign forecasting system to forecast 3 hours of the vital signs of patients in critical care using only 6 hours of previous vital sign data and evaluate the forecasted results using evaluation metrics: mean squared error (MSE), and dynamic time warping (DTW). \noindent \textbf{(2)} We evaluate the performance of three state-of-the-art forecasting models, N-BEATS, N-HiTS, and TFT, on forecasting using a publicly available dataset eICU Collaborative Research Database (eICU-CRD) \cite{pollard2018eicu}. \noindent \textbf{(3)} Since the shape and temporal changes in the vital signs are essential for healthcare providers in making clinical decisions, we use a DILATE \cite{le2019shape} loss function to capture the spatial and temporal variations in the forecasted vital signs. \textbf{(4)} We carried out a comprehensive qualitative analysis by comparing the performance of different forecasting models and assessing the performance of models on different forecasting horizons. The results of our model and experiments provide insights into data performance, potentially leading to improved patient care and outcomes in real-world scenarios.

\begin{figure}
\centerline{
\includegraphics[height=.25\textwidth]{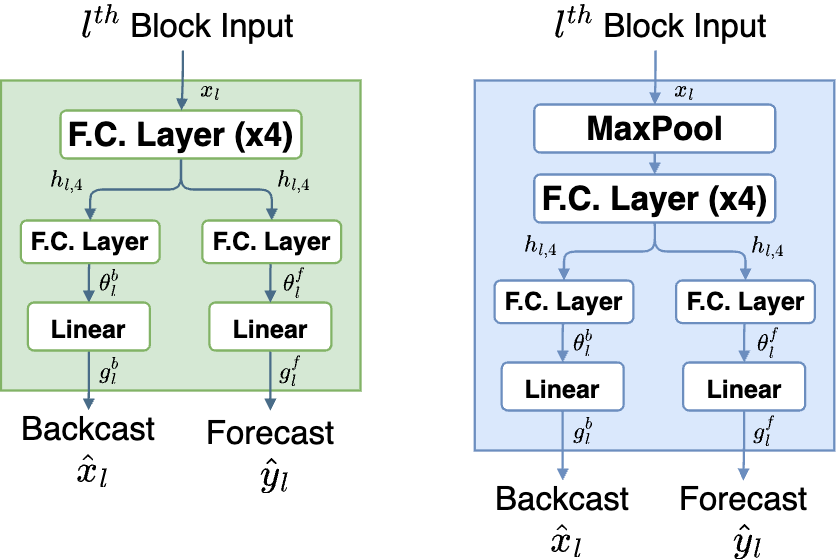}
}
\caption{The figure shows the $l^{th}$ basic block of the N-BEATS (left) and the N-HiTS architecture.}
\label{nbeats:arch}
\end{figure}

\section{Method}
\subsection{Dataset Description and Data Preprocessing} 
In our study, we employed the eICU-CRD \cite{pollard2018eicu}, focusing on the \textit{vitalPeriodic} and \textit{diagnosis} tables. We gathered 5-minute interval time series data on mean blood pressure (MBP), heart rate (HR), and respiration rate (RR) for patients with sepsis or septic shock. MBP was derived using the formula MBP = DBP + 1/3 [SBP – DBP] \cite{sainas2016mean}, where DBP and SBP are diastolic and systolic blood pressure, respectively. Missing data were imputed via forward fill, excluding cases with over 25 minutes of missing data pre-diagnosis. To impute missing vital sign data, we utilized forward fill and excluded patients with over 25 minutes of missing data before diagnosis offset. For each patient, we retained data up to 9 hours prior to diagnosis offset and arranged time series data into groups that concluded with sepsis or septic shock diagnosis. This permitted the model to recognize the pattern of vital signs that led to sepsis or septic shock diagnosis \cite{bhatti2023interpreting}. To remove outliers, we implemented a low pass filter on the vital signs and eliminated patients with less than 0.0025 standard deviations in their vital signs. The final dataset contained 4020 groups across 1442 patients. For scaling the time series data, we used min-max scaling. In the normalization of the vital signs data, we eschewed the conventional approach of using the dataset's own minimum and maximum values for scaling. Instead, we applied domain expertise to determine clinically sensible scale ranges for each vital parameter: HR from 0 to 300 bpm, MBP from 0 to 190 mmHg, and RR from 0 to 100 mmHg \cite{bhatti2023interpreting,o2020characterizing}. 

\subsection{Experiment Setup and Deep Learning Pipeline}

\begin{figure}[t]
\centerline{
\includegraphics[width=.47\textwidth]{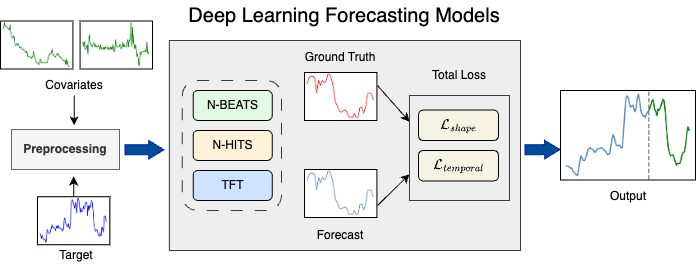}
}
\caption{The figure shows our deep learning pipeline for forecasting vital signs of patients suffering from Sepsis or Septic Shock condition in the ICUs.}
\label{fig:pipeline}
\end{figure}

Our process involves splitting the eICU-CRD dataset into training, validation, and testing groups in an 80:10:10 ratio. For the deep learning forecasting model, we used the vital signs of the training and validation groups to create a 6 hour input trajectory (72 steps) and a 3 hour future trajectory (36 steps) for prediction. We also conducted experiments with and without covariates (i.e., HR, MBP, and RR) to gauge their impact on model performance. We use standard metrics like MSE and DTW to compare the models' performance against a naive persistence model that predicted future values by repeating the last observed value in the input window for all forecasted time steps. See Figure \ref{fig:pipeline} for a visual representation of our pipeline.

\subsection{Deep Learning Forecasting Models}
We utilize state-of-the-art forecasting techniques, namely N-BEATS \cite{oreshkin2020nbeats}, N-HiTS \cite{challu2023nhits}, and TFT \cite{lim2019temporal}, to predict a 3 hour time series of vital signs: MBP and HR.

N-BEATS \cite{oreshkin2020nbeats} is a deep learning architecture designed for both univariate and multivariate time series forecasting across multiple horizons. The architecture consists of three core components, namely a basic block, a stack (a combination of blocks), and the final model. As depicted in Figure \ref{nbeats:arch}, each basic block in the architecture receives a corresponding lookback window $x_{l}$ as input and outputs two vectors, backcast ($\widehat{x}_{l}$) and forecast ($\widehat{y}_{l}$). The basic block comprises a stack of fully connected layers that generate backward ($\theta^{b}_{l}$) and forward ($\theta^{f}_{l}$) expansion coefficients (Eq. \ref{nbeats:linear}) \cite{oreshkin2020nbeats}:

\begin{equation}\label{nbeats:linear}
    % \begin{split}
    \theta^{b}_{l} = \text{Linear}^{b}_{l}(h_{l, 4}), \thickspace \thickspace
    \theta^{f}_{l} = \text{Linear}^{f}_{l}(h_{l, 4}),
    % \end{split}
\end{equation} where $h_{l, 4}$ is the output of the fourth fully connected layer in the basic block and Linear layer is a linear projection layer \cite{oreshkin2020nbeats}. And the second part consists of backward ($g^{b}_l$) and forward ($g^{f}_{l}$) basis layers that produce backcast and forecast outputs (Eq. \ref{nbeats:basis}) \cite{oreshkin2020nbeats}:

\begin{equation}\label{nbeats:basis}
    % \begin{split}
        \widehat{y}_{l} = \sum_{i=1}^{\text{dim}(\theta^{f}_{l})} \theta^{f}_{l,i}\text{v}^{f}_{l,i}, \thickspace \thickspace
        \widehat{x}_{l} = \sum_{i=1}^{\text{dim}(\theta^{b}_{l})} \theta^{b}_{l,i}\text{v}^{b}_{l,i}.
    % \end{split}
\end{equation}
Here, $v^{f}_{l,i}$ and $v^{b}_{l,i}$ are forecast and backcast basis vectors.

N-HiTS \cite{challu2023nhits}, a deep learning model tailored for long-horizon forecasting, adeptly addresses prediction volatility and computational challenges. It employs multi-rate time series sampling and a novel hierarchical interpolation method, allowing for the integration of short-term and long-term temporal effects across various time scales. The Figure \ref{nbeats:arch} highlights the architectural distinctions between N-HiTS and its counterpart, N-BEATS.
The Eq. \ref{nhits:maxpool} is used to achieve multi-rate signal sampling for an $l^{th}$ basic block \cite{challu2023nhits}.

\begin{equation}\label{nhits:maxpool}
    y^{(p)}_{t - L:t, l} = \text{MaxPool}\left(y_{t - L:t, l}, k_{l}\right),
\end{equation}
Here, $k_{l}$ is the kernel size of the MaxPool layer.

TFT \cite{lim2019temporal} is an attention-based deep learning model designed for multi-horizon time series forecasting. The architecture of TFT comprises the following building blocks: 1. Gated Residual Networks (GRNs) allow for non-linear processing to be applied only when necessary, resulting in a concentration on crucial components while suppressing extraneous ones \cite{dauphin2016language}. The GRNs are formulated based on the Eq. \ref{eqn:grn} outlined in \cite{lim2019temporal}.

\begin{equation}
    \text{GRN}_{\omega}(a, c) = \text{LayerNorm}\left(a + \text{GLU}_{\omega}{(\eta_{1})}\right)
    \label{eqn:grn}
\end{equation}
where \textit{a} and \textit{c} are the primary input and optional context vector inputs to the GRN, GLU is Gated Linear Unit \cite{dauphin2016language}, $\eta_{1} = W_{1,\omega}\eta_{2} + b_{1,\omega}$ and $\eta_{2} = \text{ELU}(W_{2,\omega}a + W_{3,\omega}c + b_{2,\omega})$ are intermediate layers, ELU is Exponential Linear Unit activation function \cite{ba2016layers}, LayerNorm is a standard layer normalization \cite{ba2016layers}, $\omega$ is an index to denote weight sharing, $W_{(.)}$ and $b_{(.)}$ are weights and biases.
2. A sequence-to-sequence encoders/decoders block that enables the identification of relationships between time steps and their surrounding values, summarizing smaller patterns. 3. A temporal multi-head attention block to identify long-term dependencies in time series data and prioritize essential patterns. 4. A variable selection network that performs instance-wise selection of variables for both static and time-dependent covariates based on the importance of their features.

% The naive persistence model provides a baseline for evaluating the performance of the other models as it does not take any external factors or trends into account. Therefore, the other models should outperform the naive persistence model if they can account for the underlying patterns in the data and make more accurate predictions.

% Ensemble not required.
% Due to the similar performance of all models, an ensemble model was created using the average model technique, where the forecasts produced by two or more models were averaged. All possible combinations of the four models (NHiTS, NBeats, TFT, and Persistence) were attempted, e.g., NHiTS-NBeats, NHiTS-NBeats-TFT, and other combinations. The ensemble combination that produced the best performance was chosen from all the combinations attempted.

\subsection{Training with DILATE Loss Function}
% Explain the narrative for using DILATE Loss and why it is important to use it here in our case.

In critical care, a patient's vital signs are crucial in determining their current and future condition. To accurately predict critical events, it's important for a deep learning forecasting model to recognize sudden changes in the time series' shape and temporal features. The loss function proposed in \cite{le2019shape} is specifically designed to address these aspects. The DILATE Loss function is composed of two distinct terms: the shape term \cite{sakoe1978dynamic,le2019shape,cuturi2017soft} and the temporal term \cite{le2019shape}, which both aim to capture changes in the time series spatial and temporal characteristics, as shown in Eq. \ref{dilate:shp_tmp} \cite{le2019shape}. 
% outlines the DILATE Loss function \cite{le2019shape}.

\begin{equation}
\begin{split}
\mathcal{L}_{\text{DILATE}}\left(\hat{\mathbf{y}}_i, \stackrel{*}{\mathbf{y}}_i\right) & = \alpha \mathcal{L}_{\text {shape }}\left(\hat{\mathbf{y}}_i, \stackrel{*}{\mathbf{y}}_i\right) \\
& +(1-\alpha) \mathcal{L}_{\text {temporal }}\left(\hat{\mathbf{y}}_i, \stackrel{*}{\mathbf{y}}_i\right)
\end{split}
\label{dilate:shp_tmp}
\end{equation}

Here, the DILATE objective function consists of the shape term ($\mathcal{L}_{shape}$) and temporal term $\mathcal{L}_{temporal}$ that compare the predictions $\hat{y}_{i}$ with the ground truth future trajectory $\stackrel{*}{ y}_{i}$. The shape and temporal terms are balanced by a hyperparameter $\alpha$ $\in$ $[0, 1]$.
% The hyperparameter $\alpha$ $\in$ $[0, 1]$ balances the shape and temporal term.

{\renewcommand{\arraystretch}{1.4}

\begin{table}[htbp]
\tiny
\caption{Performance of forecasting models on forecasting MBP and HR. Here, L1 is the MSE loss function and L2 is the DILATE loss function.}
\begin{center}
\begin{tabular}{|c|c|c|c|c|c|c|c|c|c|}
\hline
\multirow{3}{*}{\textbf{Models}} & \multirow{3}{*}{\textbf{Cov.}} & \multicolumn{4}{c|}{\textbf{Mean Blood Pressure}} & \multicolumn{4}{c|}{\textbf{Heart Rate}}\\
\cline{3-10}
& & \multicolumn{2}{c|}{\textbf{MSE*}} & \multicolumn{2}{c|}{\textbf{DTW}} & \multicolumn{2}{c|}{\textbf{MSE*}} & \multicolumn{2}{c|}{\textbf{DTW}} \\ \cline{3-10}
& & L-1 & L-2 & L-1 & L-2 & L-1 & L-2 & L-1 & L-2\\ \hline
\multirow{1}{*}{Persistence} & - & 24.55 & 24.55 & 34.50 & 34.50 & 7.35 & 7.35 & 17.52 & 17.52\\
\hline
\multirow{2}{*}{N-HiTS} & W C & \textbf{18.78} & 19.99 & 20.44 & \textbf{16.73} & \textbf{7.37} & 7.57 & 13.12 & \textbf{10.05} \\ \cline{2-10}
 & W/o C & \textbf{18.02} & 19.81 & 20.46 & \textbf{16.32} & 7.22 &  \textbf{7.18} & 13.97 &  \textbf{7.92} \\
\hline
\multirow{2}{*}{N-BEATS} & W C & \textbf{19.79} & 24.40 & 19.37 &  \textbf{18.59} & \textbf{8.73} & 10.95 &14.36 & \textbf{14.20} \\ \cline{2-10}
 & W/o C & \textbf{18.52} & 27.42 & \textbf{17.63} & 18.60 & \textbf{7.48} & 12.98 & \textbf{10.71} & 17.90 \\
\hline
\multirow{2}{*}{TFT} & W C &  \textbf{18.89} & 19.10 & 25.93 & \textbf{23.51} & 7.71 & \textbf{7.19} & 16.12 & \textbf{15.16} \\ \cline{2-10}
 & W/o C &  19.45 &  \textbf{19.00} & 25.65 & \textbf{23.46} & 8.12 &  \textbf{7.57} & 16.65 &  \textbf{15.79} \\
\hline
\end{tabular}
\label{tab:results}
\end{center}
\footnotesize{$^*$The MSE values are scaled by $1e^{-4}$ for better representation.} 
% \footnotesize{Best Ensemble for Blood Pressure: (NHiTS, TFT), \& Heart Rate: (NHiTS, TFT, Persistence)} \\
% \footnotesize{$\dagger$ Best Ensemble for Blood Pressure: (NHiTS, NBEATS) \& Heart Rate: (NHiTS, NBEATS, Persistence)} 

\end{table}}

\textbf{Shape Term:}
% Authors of DILATE loss function \cite{le2019shape} defined the shape term as Dynamic Time Warping (DTW) \cite{sakoe1978dynamic}. To make the DTW loss function differentiable, a smooth minimum operator proposed in \cite{cuturi2017soft} was used to define the differentiable shape term as \eqref{dilate:shape_term_1}. 
The DILATE loss function's shape term was based on DTW and made differentiable using a smooth minimum operator for the differentiable shape term as per \cite{le2019shape, sakoe1978dynamic, cuturi2017soft}.
Here, $\boldsymbol{\Delta}\left(\hat{y}_{i}, \stackrel{*}{y}_{i}\right)$ is the pair-wise cost matrix, the warping path is defined as a binary matrix $\mathbf{A} \subset\{0,1\}^{k \times k}$ and $\gamma > 0$. 
% $$
% \operatorname{matrix} \mathbf{A} \subset\{0,1\}^{k \times k}
% $$

\begin{equation*}
    \begin{split}
    \mathcal{L}_{\text{shape}}\left(\hat{\mathbf{y}}_i, \stackrel{*}{\mathbf{y}}_i\right) = \text{DTW}_{\gamma}\left(\hat{\mathbf{y}}_i, \stackrel{*}{\mathbf{y}}_i\right)        
    \end{split} 
\end{equation*}

\begin{equation}
    \label{dilate:shape_term_1}
    = \small{-\gamma\log\left(\sum_{\mathbf{A} \in \mathcal{A}_{k,k}} \exp \left(-\frac{\left\langle\mathbf{A}, \boldsymbol{\Delta}\left(\hat{\mathbf{y}}_i, \stackrel{*}{\mathbf{y}}_i\right)\right\rangle}{\gamma}\right)\right)}
\end{equation}

\textbf{Temporal Term:}
The temporal loss function is inspired from the time distortion index \cite{frias2017assessing,vallance2017towards}, where $Z = \sum_{\mathbf{A} \in \mathcal{A}_{k, k}} \exp^{-\frac{\langle\mathbf{A}, \boldsymbol{\Delta}(\hat{\mathbf{y}}_i, \stackrel{*}{\boldsymbol{y}}_i)\rangle}{\gamma}}$ and $\mathbf{\Omega}$ is a square matrix of size $k \times k$ penalizing each element $\hat{y}^{h}_{i}$ being associated to an $\stackrel{*}{y}^{j}_{i}$, for $h \neq j$. Similar to \cite{le2019shape}, we have used the squared penalization for our experiments, i.e., $\Omega(h,j) = \frac{1}{k^2}(h - j)^2$. The differentiable loss used for the temporal term is given as: 

\begin{equation}
\begin{split}
\mathcal{L}_{\text {temporal}}\left(\hat{\mathbf{y}}_i, \stackrel{*}{\mathbf{y}}_i\right) & := \left\langle\mathbf{A}_\gamma^*, \boldsymbol{\Omega}\right\rangle \\
& = \frac{1}{Z} \sum_{\mathbf{A} \in \mathcal{A}_{k, k}}\langle\mathbf{A}, \boldsymbol{\Omega}\rangle \exp ^{-\frac{\left\langle\mathbf{A}, \boldsymbol{\Delta}\left(\hat{\mathbf{y}}_i, \stackrel{*}{\boldsymbol{y}}_i\right)\right\rangle}{\gamma}}
\end{split}
\end{equation}

\section{Results and Discussion}

\begin{figure}
     \centering  
     \begin{subfigure}[b]{0.3\textwidth}
         \centering
         \includegraphics[width=\textwidth]{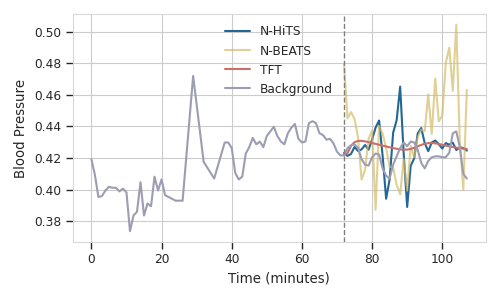}
         \caption{Forecasting results}
         \label{fig:forcasting}
     \end{subfigure}
     % \hfill
     \begin{subfigure}[b]{0.18\textwidth}
         \centering
         \includegraphics[width=\textwidth]{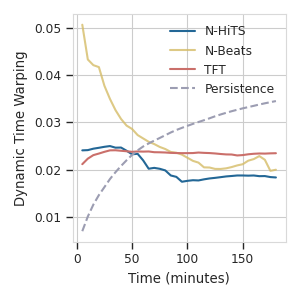}
         \caption{Error analysis}
         \label{fig:error}
     \end{subfigure}
        \caption{N-HiTS, N-BEATS and TFT without covariates using DILATE loss function.}
        \label{fig:results}
\end{figure}

We evaluate the performance of N-HiTS, N-BEATS, and TFT using test samples, summarized in Table \ref{tab:results} through MSE and DTW error scores. Figure \ref{fig:forcasting} indicates that TFT captures the trends well, while N-HiTS better retains fluctuations within a certain range. Interestingly, N-HiTS performs better without covariates, whereas TFT benefits from them. 
% We particularly highlight the promising results of N-HiTS without covariates when evaluated with DTW, meriting further exploration.
Our evaluation of trained models assesses their performance over a range of forecasting horizons, from 5 minutes to 3 hours, using the DTW metric. To accomplish this, we generate horizon windows of increasing length, starting at one timestep (5 minutes) and ending with 36 timesteps (180 minutes). This enables a thorough evaluation of each model's performance compared to the Persistent model. Figure \ref{fig:error} displays the DTW errors for the models N-HiTS, N-BEATS, TFT, and Persistence for MBP. 
Deep learning models initially have higher DTW errors than the persistence model near the 50-minute mark but later demonstrate a downward trend and eventually outperform the persistence model, a pattern anticipated due to the gradual change in vital signs that tend to stay near prior values.

\section{Conclusion and Future Work}
Our research has introduced a deep learning system that accurately forecasts vital signs in ICU patients with sepsis. We tested three sophisticated models and applied a DILATE loss function to capture the vital sign fluctuations crucial for timely clinical intervention. The promising results indicate that this approach could greatly assist in the early detection of conditions like septic shock. However, future efforts should look into integrating more clinical parameters and multimodal data to enhance forecasting accuracy. This refinement and broader data incorporation have the potential to further improve the system's utility in real-world clinical settings, ultimately benefiting patient care in critical situations.

\bibliographystyle{IEEEtran}

\bibliography{reference}

% \color{red}
% IEEE conference templates contain guidance text for composing and formatting conference papers. Please ensure that all template text is removed from your conference paper prior to submission to the conference. Failure to remove the template text from your paper may result in your paper not being published.

\end{document}